\documentclass{article}
\usepackage{ijcai26}

\usepackage{times}
\usepackage{soul}
\usepackage{url}
\usepackage[hidelinks]{hyperref}
\usepackage[utf8]{inputenc}
\usepackage[small]{caption}
\usepackage{graphicx}
\usepackage{amsmath}
\usepackage{amsthm}
\usepackage{amssymb}
\usepackage{booktabs}
\usepackage{algorithm}
\usepackage{algorithmic}
\usepackage{subcaption}
\usepackage{multirow}
\usepackage{cleveref}
\usepackage{tikz}
\usepackage{bm}
\usepackage{makecell}
\usepackage{newfloat}
\usepackage{listings}
\usepackage{newfloat}
\usepackage{listings}
\title{Belief-Contraction-Driven Active Inverse Source Localization and Characterization}

\author{
Yiwei Shi$^1$\and
Mengyue Yang$^{1}$\thanks{Corresponding author.}\and
Qi Zhang$^2$\footnotemark[1]\and
Cunjia Liu$^3$\footnotemark[1]\and
Weinan Zhang$^4$\footnotemark[1]\And
Weiru Liu$^{1}$\footnotemark[1]
\affiliations
$^1$School of Engineering Mathematics and Technology, University of Bristol\\
$^2$Department of Computer Science and Technology, Tongji University\\
$^3$Department of Aeronautical and Automotive Engineering, Loughborough University\\
$^4$School of Computer Science, Shanghai Jiao Tong University\\
\emails
yiwei.shi@bristol.ac.uk
}

\begin{document}

\maketitle

\begin{abstract}
Active inverse source localization and characterization (ISLC) in dynamic fields requires sequential decision making under partial observability, where a mobile sensor must infer latent source parameters from sparse, noisy readings. We introduce a belief-contraction-driven approach that unifies inference, stopping, and control. An attention-augmented particle filter stabilizes Bayesian belief updates through ESS-based resampling, feature-aware sparse attention smoothing, and Metropolis–Hastings rejuvenation that preserves the filtering posterior. Belief contraction (posterior dispersion) defines both a termination rule and a goal-aligned intrinsic reward, enabling reinforcement learning without distance-to-source shaping. Across seven field modalities, spatial out-of-distribution tests, and nonstationary source shifts, our agent (ATT-PFRL) achieves higher completion, faster convergence, and more accurate localization than planning and RL+Bayes baselines under similar computation. Fixed-trajectory studies also show improved ESS and lower RMSE, isolating the benefit of the inference layer.

\end{abstract}

\addtocontents{toc}{\protect\setcounter{tocdepth}{-1}}

\section{Introduction}

Accurately and rapidly localizing unknown sources in spatially distributed fields, such as gas leaks, pollutant releases, or electromagnetic anomalies,is a critical capability for industrial safety, environmental monitoring, and emergency response. These tasks are commonly formulated as \textbf{Inverse Source Localization and Characterization (ISLC)}, where the objective is to infer latent source and environmental parameters (e.g., source position, intensity, and transport conditions) from \textit{sparse, noisy, and localized measurements} collected by a mobile agent \cite{steiner2001large}. Active ISLC is inherently a \textbf{partially observable} sequential decision-making problem. Observations are local and often weakly informative, while the latent parameters remain unobserved; thus, effective behavior must be conditioned on a \textbf{belief} over those parameters rather than on raw measurements alone. This setting creates three practical challenges. First, \textit{intermediate feedback is sparse and implicit}: the environment typically does not provide dense rewards that indicate progress toward successful localization, making reinforcement learning (RL) unstable or heavily dependent on hand-crafted shaping. Second, maintaining a reliable posterior under sparse/noisy sensing is non-trivial: standard particle filters can suffer from \textit{weight degeneracy} and resampling-induced \textit{particle impoverishment}, leading to overconfident yet unreliable beliefs. Third, real deployments demand robustness across varying field modalities and distribution shifts (e.g., changed wind or spatial regimes), where reactive heuristics or narrowly trained policies may fail to generalize.

\begin{figure}[!t]
\centering
% 第一行
\includegraphics[width=\linewidth]{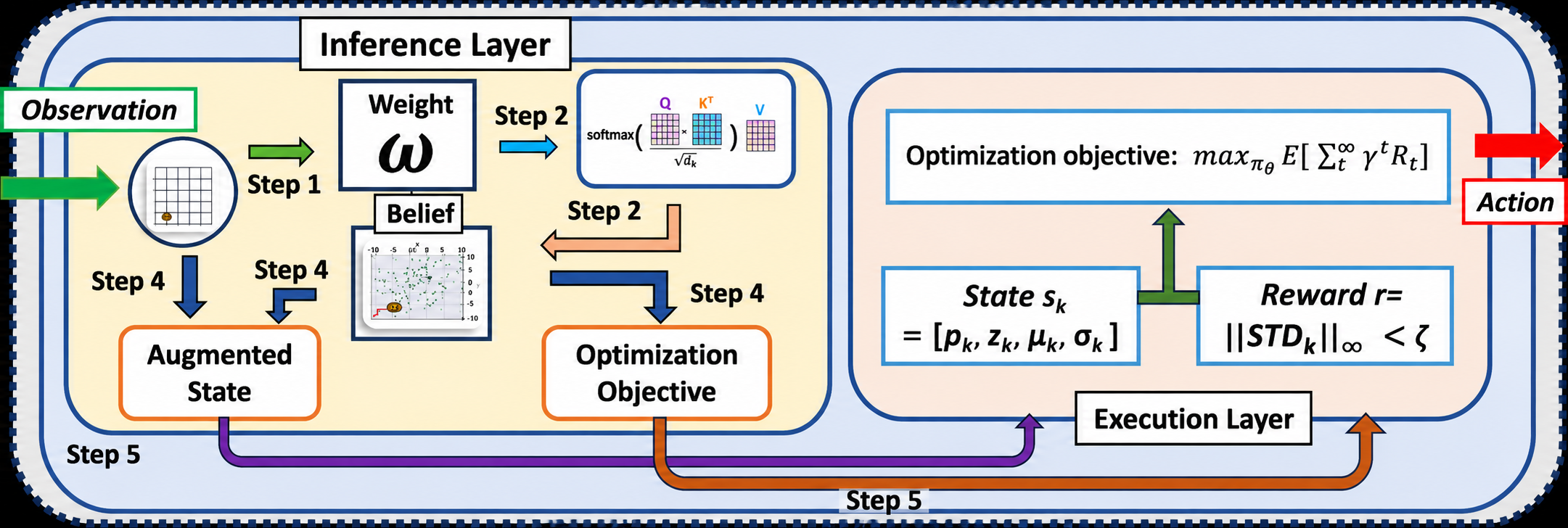}%
\caption{Attention-Based Inference Framework}
\label{fig_framework}
\end{figure}
Prior work approaches ISLC through Bayesian planning and information-theoretic search \cite{vergassola2007infotaxis,hutchinson2018information}, bio-inspired plume tracing and casting strategies, or deep RL methods for adaptive exploration \cite{zhao2022deep,hu2019plume,park2022source}. However, existing solutions often face a trade-off between (i) \textit{principled inference} and (ii) \textit{goal-aligned control}. Information-theoretic planning can be computationally demanding and sensitive under partial observability, while RL approaches frequently rely on reward shaping that may be misaligned with the true objective (pinpointing the source) and can generalize poorly across conditions. 

In this paper, we propose a \textit{belief-contraction-driven framework} that couples Bayesian belief maintenance with belief-conditioned decision making for active ISLC. Our key idea is to treat \textit{posterior contraction}—the reduction of uncertainty in the inferred belief—as the common objective that links inference, termination, and learning. Concretely, we (i) maintain a particle-based posterior over latent parameters and (ii) use a contraction criterion (based on posterior dispersion) as a \textit{task-completion signal} and as a \textit{sparse intrinsic reward} for learning. This converts implicit “mission completion” into an explicit, goal-aligned learning signal, avoiding distance-to-source shaping and directly reflecting whether the agent has localized the source with sufficient confidence.

To make belief contraction meaningful and robust under sparse/noisy sensing, we introduce an \textit{attention-augmented particle inference layer} that mitigates degeneracy while controlling computational overhead. Starting from sequential importance sampling, we stabilize belief updates via (1) \textit{ESS-triggered resampling} when weight collapse is detected, (2) \textit{feature-aware sparse attention smoothing} that redistributes weight mass among feature-similar hypotheses as a controlled perturbation on the simplex, and (3) \textit{MCMC rejuvenation with Metropolis–Hastings correction} to restore diversity while preserving posterior invariance. The resulting belief state (posterior moments and uncertainty summaries) serves as the input to an execution policy. In this work, we instantiate the executor as an RL agent,\textit{ATT-PFRL},that learns exploration strategies conditioned on the belief, optimizing the sparse contraction-based intrinsic reward until the termination criterion is met. We validate the proposed framework in a suite of ISLC environments spanning diverse field modalities and challenging shifts. We further evaluate \textit{out-of-distribution spatial shifts} and \textit{nonstationary source changes}, highlighting robustness to distribution mismatch and dynamic regimes. To isolate inference improvements from policy-induced data collection, we additionally conduct \textit{fixed-trajectory evaluations} that compare inference variants under identical observation sequences, and we perform dedicated analyses of \textit{cessation threshold sensitivity} and \textit{intrinsic reward design}, verifying that the contraction signal is calibrated against ground-truth localization quality.

Our main contributions are: 1.) A unified framework that aligns inference, termination, and learning by using posterior contraction as both a stopping criterion and a sparse intrinsic reward for belief-conditioned control.
2.) An efficient degeneracy-mitigation pipeline combining ESS-triggered resampling, feature-aware sparse attention smoothing, and MH-corrected rejuvenation to stabilize belief updates under sparse/noisy observations while preserving posterior correctness.
3.) Extensive experiments across diverse field types, out-of-distribution regimes, and nonstationary shifts, including inference decoupling and reward/cessation validations that directly support the proposed design choices.

\section{Related Work}
\textbf{Traditional planning-based methods} for ISLC typically fall into two categories: Information-Theoretic Approaches and Bio-Inspired Methods. The former strategies aim to reduce uncertainty about the source location, with \textit{Infotaxis} \cite{vergassola2007infotaxis} minimizing posterior variance, \textit{Entrotaxis} \cite{hutchinson2018entrotaxis} focusing on high-entropy regions, and \textit{DCEE} \cite{chen2021dual} balancing exploration and exploitation through uncertainty-driven decision-making. Although these methods have proven effective in controlled settings, they often rely on strong assumptions, leading to reduced efficiency and success rates when conditions deviate from their underlying assumptions. Meanwhile, the latter algorithms  like \textit{plume-tracing} \cite{farrell2005chemical} and \textit{zigzag} \cite{balkovsky2002olfactory,lochmatter2008comparison} mimic the \textit{gradient-following behaviors} used by insects and animals—detecting changes in chemical concentration and adjusting movement to reacquire the plume if it is lost. Though effective in stable environments, these reactive approaches often fail or converge slowly in nonstationary, noisy, or complex fields, where their assumptions and simple local cues become inadequate. Unlike planning methods, \textbf{RL} \cite{mnih2015human,schulman2017proximal,lillicrap2015continuous,hu2025pmat,shi2026distill,shi2025dynamic} excels in high-dimensional tasks but faces two major hurdles in ISLC: sparse or implicit rewards, where locating a hidden source offers little immediate feedback, and inference, as RL inherently lacks the capacity to directly deduce hidden variables in noisy, evolving fields. Recent efforts combine RL with Bayesian updates to address this, but many approaches still rely on hand-crafted rewards or are domain-specific. PC-DQN \cite{zhao2022deep} combines particle clustering and deep Q-networks for feature extraction, demonstrating high success rates in turbulent environments. LSTM-based RL \cite{hu2019plume} enhances historical trajectory encoding for robust plume tracing under dynamic conditions, while DDPG with GMM \cite{park2022source} features facilitates real-time, efficient source estimation in stochastic scenarios. Among these methods, AGDC \cite{shi2025autonomous} excels in handling sparse-feedback settings by detecting and halting upon goal completion while leveraging inference to improve decision-making under uncertainty. Its three variants extend its functionality: AGDC-KLD minimizes KL divergence for accurate localization \cite{filippi2010optimism}, AGDC-ENT focuses on high-uncertainty areas\cite{haarnoja2018soft}, and AGDC-EE balances exploration and exploitation \cite{li2022concurrent}. Despite its advancements, AGDC still faces challenges in efficiency, particularly in inference speed and scalability, for dynamic real-time environments.

\section{Preliminaries}
\label{sec:Peliminaries}

\subsection{Convection-Diffusion Equation}

Many seemingly disparate natural phenomena, such as \textit{pollutant dispersion}, \textit{gas diffusion}, and \textit{electric field distributions}, share common physical mechanisms: \textbf{diffusion}, \textbf{convection}, and \textbf{external sources}. These mechanisms can be mathematically described by the general \textit{convection-diffusion equation} (CDE) \cite{holley1969unified}, which provides a versatile framework:
\begin{equation}
\alpha \nabla^2 \phi - \vec{v}\cdot\nabla\phi + k_r \phi + S(x,y) = 0,
\end{equation}
where $\phi(x, y)$ denotes the field variable (e.g., concentration or temperature), with $\alpha\nabla^2\phi$, $-\vec{v}\cdot\nabla\phi$, and $S(x, y)$ accounting for diffusion, convection, and external sources, respectively. The parameters, diffusion coefficient $\alpha$, convection velocity $\vec{v}$, reaction rate $k_r$, and source term $S(x,y)$, can be adapted to describe various physical phenomena, from heat conduction to pollutant dispersion.

\subsection{Gaussian Plume Model}
The Gaussian plume model is derived as a steady-state analytical solution of the CDE, balancing simplicity and computational efficiency:
$ \phi(x, y) = \frac{q_s}{4 \pi \alpha \|\boldsymbol{p}-\boldsymbol{p}_s\|} \exp\left(-\frac{\|\boldsymbol{p}-\boldsymbol{p}_s\|}{\lambda} - \frac{(x - x_s)u_x + (y - y_s)u_y}{2 \alpha}\right)$, where $q_s$ is the source strength, $\boldsymbol{p} = (x,y)$ and $\boldsymbol{p}_s = (x_s,y_s)$ are the coordinates of the observation point and source, respectively, $u_x,u_y$ are the components of the convection velocity $\vec{v}$, $\lambda=\sqrt{\gamma\tau_s\Big/\left(1+{u_s^2\tau_s}/{4\gamma}\right)}$  is a decay parameter,   \(\tau_s\) is the time duration and $\alpha$ is the diffusion coefficient consistent with the general CDE notation. 
When wind is parameterized by speed and direction, we use $u_x=u_s\cos\varphi_s$ and $u_y=u_s\sin\varphi_s$.
We parameterize wind as $(u_x,u_y)$.

\subsection{Partially Observable Markov Decision Process}

Dynamic field modeling frequently encounters uncertainties arising from incomplete state information and noisy observations. These challenges can be systematically addressed through the \textit{Partially Observable Markov Decision Process (POMDP)}, defined by the tuple $(S, \Omega, A, T, O, R, \gamma)$. At each discrete time step, the agent receives an observation $o_t \in \Omega$ according to $O(o_t|s_t,a_{t-1})$, takes an action $a_t\in A$, and the environment transitions to $s_{t+1}$ via $T(s_{t+1}|s_t,a_t)$. The objective is to find a policy that maximizes the expected discounted return. Under partial observability, the policy conditions on a belief state. We model active ISLC as a POMDP where the \emph{full} system state is
$s_t = (\boldsymbol{p}_t, \Theta_t)$, consisting of the agent position $\boldsymbol{p}_t\in\mathbb{R}^2$ and latent source/environment parameters
$\Theta_t$ (e.g., source location/strength, wind, diffusivity).
The action is a discrete motion command $a_t \in \{\uparrow,\downarrow,\leftarrow,\rightarrow\}$ that deterministically updates the position
$\boldsymbol{p}_{t+1}=f(\boldsymbol{p}_t,a_t)$ (with boundary clipping).
The observation is $o_t=(\boldsymbol{p}_t, z_t)$, where $z_t$ is the noisy sensor reading at $\boldsymbol{p}_t$ with likelihood $p(z_t \mid \boldsymbol{p}_t,\Theta_t)$.
We consider both (i) stationary settings where $\Theta_{t+1}=\Theta_t$ and (ii) nonstationary settings with a piecewise-smooth transition
$p(\Theta_{t+1}\mid \Theta_t)$ (e.g., wind/source changes).
Since $\Theta_t$ is unobserved, the policy conditions on the belief
$b_t(\Theta)=p(\Theta_t\mid o_{1:t})$ maintained by particle filtering.
Finally, instead of assuming dense external rewards, we define an \emph{intrinsic termination reward} based on posterior uncertainty, aligning learning/planning with the ISLC objective. \textcolor{black}{Objective: Therefore, our objective is to infer the posterior distribution $p(\Theta \mid o_{1:t})$ of the field-model parameters$
\Theta = [x_s, y_s, q_s, u_x, u_y, \alpha, \lambda]^\top$
from sequential observations $o_{1:t}$ and to use this belief to guide action selection.}

\section{Methodology}
\label{sec:methodology}

\paragraph{Overview.} ISLC is formulated as a POMDP in Sec.~\ref{sec:Peliminaries}.
We propose a belief-driven decision framework with the following loop:
\textit{(i) maintain a Bayesian belief over unknown source parameters using a particle approximation;}
\textit{(ii) turn belief contraction into a sparse intrinsic learning signal;}
\textit{(iii) learn a policy that actively queries informative locations.}
In this paper, we present the RL-based execution strategy, \textbf{ATT-PFRL}, as the proposed method.
Planning-style baselines are \textbf{evaluated} only in Sec.~\ref{sec:experiments}.

\subsection{Bayesian Belief Approximation via SIS}
\label{sec:bayes_pf}

\textbf{State of uncertainty (belief).}
At step $k$, the agent observes
$o_k = (\boldsymbol{p}_k, z_k)$,
where $\boldsymbol{p}_k\in\mathbb{R}^2$ is the agent position and $z_k\in\mathbb{R}$ is the sensor reading.
Let $\Theta$ denote the latent (unknown) source parameters (e.g., source location and emission strength; the exact parameterization
is defined in Sec.~\ref{sec:Peliminaries}).
We maintain the \emph{filtering posterior} (belief) $b_k(\Theta) := p(\Theta \mid o_{1:k})$, where $o_{1:k}$ is the observation history up to step $k$. The likelihood can be highly non-linear and multi-modal; keeping a posterior (instead of a single estimate) provides uncertainty information that is useful for termination and for belief-conditioned control. \textbf{Particle approximation.}
We approximate $b_k$ using $N_k$ weighted particles
$\{(\Theta_k^i, w_k^i)\}_{i=1}^{N_k}, w_k^i\ge 0,\ \sum_{i=1}^{N_k} w_k^i = 1$, via $ b_k(\Theta)\approx \sum_{i=1}^{N_k} w_k^i\,\delta(\Theta-\Theta_k^i)$,
where $\delta(\cdot)$ is the Dirac delta.
Here, $\Theta_k^i$ is a hypothesis of the source parameters and $w_k^i$ is its posterior mass.
Unless otherwise stated, $N_k$ can be fixed; we later allow adaptive $N_k$ for efficiency. \textbf{Sequential importance sampling (SIS).}
The filtering posterior satisfies
$ p(\Theta_{k+1}\mid o_{1:k+1})
\propto
p(o_{k+1}\mid \Theta_{k+1})\,p(\Theta_{k+1}\mid\Theta_k)\,p(\Theta_k\mid o_{1:k})$, where $p(o_{k+1}\mid \Theta_{k+1})$ is the likelihood and $p(\Theta_{k+1}\mid\Theta_k)$ is a (possibly trivial) parameter transition model.
Given a proposal distribution $q(\Theta_{k+1}\mid \Theta_k,o_{1:k+1})$, SIS updates the weights by
\begin{align}
\bar{w}_{k+1}^{i}
~\propto~
w_{k}^{i}
\,\times\,
\frac{
   p\bigl(o_{k+1}\mid\Theta_{k+1}^{i}\bigr)\,
   p\bigl(\Theta_{k+1}^{i}\mid\Theta_{k}^{i}\bigr)
}{
   q\bigl(\Theta_{k+1}^{i}\mid\Theta_{k}^{i}, o_{1:k+1}\bigr)
},
\label{eq:SIS-weight}
\end{align}
followed by normalization $w_{k+1}^i=\bar{w}_{k+1}^i/\sum_{j=1}^{N_k} \bar{w}_{k+1}^j$. This correction accounts for sampling from $q$ rather than directly from the posterior.

\textbf{Stationary parameters (default in ISLC).} In many ISLC episodes, $\Theta$ is stationary within an episode. We set
$\Theta_{k+1}^i=\Theta_k^i,p(\Theta_{k+1}\mid \Theta_k)=\delta(\Theta_{k+1}-\Theta_k)$, and choose $q(\Theta_{k+1}^i\mid \Theta_k^i,o_{1:k+1})=\delta(\Theta_{k+1}^i-\Theta_k^i)$.
Then \eqref{eq:SIS-weight} reduces to pure Bayesian reweighting: $\bar{w}_{k+1}^{i} = w_{k}^{i}\cdot p\bigl(o_{k+1}\mid \Theta_{k}^{i}\bigr)$. This is the simplest and most efficient update when parameters do not drift within an episode. \textbf{Belief moments used downstream.}
We summarize $b_k$ using the weighted mean and covariance:$\boldsymbol{\mu}_k = \sum_{i=1}^{N_k} w_k^i\Theta_k^i,
\boldsymbol{\Sigma}_k=\sum_{i=1}^{N_k} w_k^i(\Theta_k^i-\boldsymbol{\mu}_k)(\Theta_k^i-\boldsymbol{\mu}_k)^\top$. We also define a compact uncertainty vector
$ \mathrm{STD}_k=\sqrt{\mathrm{diag}(\boldsymbol{\Sigma}_k)}$.
$\boldsymbol{\mu}_k$ provides a point summary; $\boldsymbol{\Sigma}_k$ (and $\mathrm{STD}_k$) quantifies uncertainty for cessation and for belief-conditioned control. \textbf{Convergence and finite-particle issues.} Under standard assumptions (e.g., bounded likelihood and proposal support coverage), SIS converges to the true posterior as $N_k\to\infty$.
In practice, with finite $N_k$, weights may become highly concentrated (degeneracy), and resampling may reduce diversity.
This motivates the degeneracy mitigation block in Sec.~\ref{sec:att_mcmc}.

\subsection{Degeneracy Mitigation}
\label{sec:att_mcmc}

\paragraph{Goal of this block.}
The SIS update is repeated at every decision step; thus we need a stabilizing mechanism that (i) reduces weight degeneracy, (ii) preserves
useful particle diversity, and (iii) keeps per-step overhead manageable.
We implement a unified \emph{stabilize--accelerate} block consisting of:
\textcolor{black}{\textbf{(1)} ESS-triggered resampling},
\textcolor{black}{\textbf{(2)} feature-aware \emph{sparse} attention smoothing as a controlled perturbation on weights},
and \textcolor{black}{\textbf{(3)} MH-corrected MCMC rejuvenation whose Markov kernel leaves the filtering posterior invariant}.
We also allow the particle budget $N_k$ to be adapted in later stages for efficiency.

\noindent \textcolor{black}{\textbf{(1) ESS-triggered resampling.}}
Given normalized weights $\{w_k^i\}_{i=1}^{N_k}$, we compute effective sample size
$ \mathrm{ESS}_k \;=\; \frac{1}{\sum_{i=1}^{N_k} (w_k^i)^2}$.
If $\mathrm{ESS}_k < \eta N_k$, where $\eta\in(0,1)$ is a threshold ratio, we resample (systematic or multinomial) and reset weights to $1/N_k$.
ESS quantifies the degree of degeneracy; resampling re-allocates particles toward high-likelihood regions when only a few particles carry most of the mass.

\noindent \textcolor{black}{\textbf{(2) Feature-aware sparse attention smoothing.}}
Resampling alone can reduce diversity, and using scalar weights as $Q{=}K{=}V{=}w$ ignores particle content.
We therefore compute attention using particle features.
For each particle $i$, define a feature vector
$\boldsymbol{f}^i_k =  \left[\Theta^i_k;\ \log p(o_k\mid \Theta^i_k);\ \log(w^i_k+\epsilon_{\mathrm{num}})\right]$, where $\epsilon_{\mathrm{num}}>0$ is a small constant for numerical stability.
We embed it via a shared projection $g(\cdot)$ (e.g., an MLP):
$\boldsymbol{e}^i_k=g(\boldsymbol{f}^i_k)\in\mathbb{R}^d$, where $d$ is the embedding dimension.

\noindent\textbf{Sparse attention construction.} For each particle $i$, we build a neighbor set $\mathcal{S}_i$ using approximate $m$-nearest neighbors (mNN) in the embedding space, where $m$ is a sparsity budget. We compute softmax attention only over $\mathcal{S}_i$:
$\alpha_{ij}^{(s)} \propto \exp\left((\boldsymbol{e}^i_k)^\top \boldsymbol{e}^j_k/\sqrt{d}\right)
\ \text{s.t.}\ j\in\mathcal{S}_i,
\alpha_{ij}^{(s)}=0\ \text{if}\ j\notin\mathcal{S}_i$,
and $\sum_{j=1}^{N_k}\alpha_{ij}^{(s)}=1$.
We optionally expand $\mathcal{S}_i$ until the retained probability mass exceeds $1-\delta$, where $\delta\in(0,1)$ bounds the dropped tail mass per row.
Let $A_k^{(s)}\in\mathbb{R}^{N_k\times N_k}$ denote the resulting row-stochastic attention matrix with $(A_k^{(s)})_{ij}=\alpha_{ij}^{(s)}$.

\noindent\textbf{Attention smoothing on the simplex.}
We apply a convex smoothing operator on weights:$\tilde{\boldsymbol{w}}_k
= (1-\varepsilon)\boldsymbol{w}_k
+\varepsilon \big(A_k^{(s)}\big)^\top \boldsymbol{w}_k, \varepsilon\in[0,1]$,
followed by normalization. The update mixes weight mass along feature-similar particles; this can reduce extreme concentration while remaining a controlled perturbation of the original weights.

\noindent\textbf{Proposition 1 (Simplex preservation).}
If $\boldsymbol{w}_k$ lies on the probability simplex and $A_k^{(s)}$ is row-stochastic, then $\tilde{\boldsymbol{w}}_k$ lies on the
probability simplex.

\noindent\textbf{Proposition 2 (Controlled sparse approximation).}
If the dropped tail mass per row is at most $\delta$, then the deviation between dense and sparse smoothing satisfies
$\|\tilde{\boldsymbol{w}}_k^{\mathrm{dense}}-\tilde{\boldsymbol{w}}_k^{\mathrm{sparse}}\|_1 \le 2\varepsilon\delta$
(see App. for details).

\noindent \textcolor{black}{\textbf{(3) MH-corrected MCMC rejuvenation.}}
Using $\tilde{\boldsymbol{w}}_k$, we compute $(\boldsymbol{\mu}_k,\boldsymbol{\Sigma}_k)$ with $\boldsymbol{\Sigma}_k\leftarrow \boldsymbol{\Sigma}_k+\epsilon_{\mathrm{num}}\mathbf{I}$ for numerical stability.
To mitigate resampling-induced impoverishment while keeping the update statistically principled, we apply an MH-corrected \emph{move} step
that targets the filtering posterior $\pi_k(\Theta)=p(\Theta\mid o_{1:k}) \propto p(\Theta)\prod_{t=1}^{k}p(o_t\mid \Theta)$. In implementation, rejuvenation is invoked when degeneracy handling is active (e.g., after ESS-triggered resampling), and we run
$N_{ESS}$ MH moves per particle, where $N_{ESS}$ is a small fixed integer.
Resampling duplicates particles; MH moves can restore diversity while maintaining the target posterior as the stationary distribution. \textbf{Target evaluation.}
We evaluate the MH ratio in log-space using the unnormalized log-density $
\log \tilde{\pi}_k(\Theta)=\log p(\Theta)+\sum_{t=1}^{k}\log p(o_t\mid \Theta)$, by replaying the stored within-episode history $o_{1:k}$; thus each proposal requires $O(k)$ likelihood evaluations and the
total overhead is $O(N_k N_{ESS})$ per rejuvenation call. \textbf{Proposals.}
We use a Gaussian random-walk proposal
$\Theta^i_{\mathrm{new}}
= \Theta^i_k + h_{\mathrm{rw}}\,\boldsymbol{\Sigma}_k^{1/2}\boldsymbol{\xi},
\boldsymbol{\xi}\sim\mathcal{N}(0,\mathbf{I})$,
where $h_{\mathrm{rw}}>0$ is a step size.
When $\nabla_\Theta \log p(o_t\mid \Theta)$ is available, we also consider a Langevin-style proposal with an attention-weighted direction
$ x_k^i=\sum_{j=1}^{N_k} \alpha^{(s)}_{ij}\,\nabla_\Theta \log p(o_k\mid \Theta^j_k)$,
$\Theta^i_{\mathrm{new}} = \Theta^i_k + \tfrac{h^2}{2}\,\boldsymbol{\Sigma}_k\, x_k^i + h\,\boldsymbol{\Sigma}_k^{1/2}\boldsymbol{\xi},
\boldsymbol{\xi}\sim\mathcal{N}(0,\mathbf{I})$,
where $h>0$ is a step size.
We monitor the MH acceptance rate and tune $(h_{\mathrm{rw}},h)$ to avoid degenerate accept/reject behavior. \textbf{MH correction.}
Each proposal is accepted with probability
$
a\big(\Theta^i_k\!\rightarrow\!\Theta^i_{\mathrm{new}}\big)
= \min\!\left\{
1,\,
\frac{\tilde{\pi}_k(\Theta^i_{\mathrm{new}})\,q(\Theta^i_k\mid \Theta^i_{\mathrm{new}})}{\tilde{\pi}_k(\Theta^i_k)\,q(\Theta^i_{\mathrm{new}}\mid \Theta^i_k)}
\right\}$, where $q(\cdot\mid\cdot)$ is the proposal density. The resulting Markov kernel leaves $\pi_k$ invariant.

\noindent\textbf{Proposition 3 (Posterior invariance).}
Assuming exact evaluation of $\tilde{\pi}_k(\Theta)$ up to a normalizing constant, the MH acceptance rule above enforces
detailed balance with respect to $\pi_k(\Theta)=p(\Theta\mid o_{1:k})$; therefore, the rejuvenation kernel preserves posterior invariance.

% \paragraph{Complexity and interface.}
% With sparse neighbors of average size $\bar m\ll N_k$, attention smoothing costs $O(N_k\,\bar m\,d)$ per step (instead of $O(N_k^2 d)$ when dense).
% The belief moments $(\boldsymbol{\mu}_k,\boldsymbol{\Sigma}_k)$ and uncertainty $\mathrm{STD}_k$ are passed to the cessation rule and policy module below.

\subsection{Cessation and Belief-to-Policy Learning}
\label{sec:cessation}

\textbf{Cessation and intrinsic reward.}
We stop an episode when posterior uncertainty is sufficiently small. We compute $ \mathrm{STD}_k = \sqrt{\mathrm{diag}(\boldsymbol{\Sigma}_k)}$. Termination occurs when $\|\mathrm{STD}_k\|_\infty < \zeta$, where $\zeta>0$ is a user-defined threshold. Because \textsc{ISLCenv} does not provide shaped external rewards, we define a sparse, goal-aligned intrinsic signal:
$r_{k+1}=\mathbb{I}\left[\|\mathrm{STD}_{k+1}\|_\infty < \zeta\right]$. 
This reward directly reflects the ISLC objective (localize by reducing posterior uncertainty) without introducing additional hand-crafted shaping terms. \textbf{Belief-conditioned control.}
We define a belief-augmented state for control:
$s_k=[\boldsymbol{p}_k, z_k, \boldsymbol{\mu}_k, \boldsymbol{\sigma}_k], \boldsymbol{\sigma}_k=\sqrt{\mathrm{diag}(\boldsymbol{\Sigma}_k)}$. ATT-PFRL learns a stochastic policy $\pi_\theta(a_k\mid s_k)$ that maximizes the discounted return
$\max_{\pi_\theta}\ \mathbb{E}\left[\sum_{t=0}^{\infty}\gamma_{RL}^t r_{k+t+1}\right]$,
where $\gamma_{RL}\in(0,1)$ is the discount factor.
We employ an actor-critic algorithm for discrete actions: the critic estimates $V_\psi(s_k)$ and the TD error is
$\delta_k = r_{k+1} + \gamma_{RL} V_\psi(s_{k+1}) - V_\psi(s_k)$, with standard gradient updates for $\theta$ and $\psi$.
Code is in Alg~\ref{alg:att_pfrl_core}.

\begin{algorithm}[htbp]
\caption{ATT-PFRL Core Loop (Inference + Execution)}
\label{alg:att_pfrl_core}
\begin{algorithmic}[1]
\STATE Initialize agent position \(p_0\), step \(k \leftarrow 0\)
\STATE Initialize particles \(\{(\Theta_i^{0}, w_i^{0})\}_{i=1}^{N} \sim p(\Theta)\), \(w_i^{0} \leftarrow 1/N\)
\WHILE{$k < K_{\max}$ \AND $\|\sigma_k\|_\infty \ge \zeta$}
  \STATE Observe $o_k=(p_k,z_k)$
  \FOR{$i=1,\dots,N$}
    \STATE $\Theta_i^{k}\leftarrow \Theta_i^{k-1}$ \; (or $\sim p(\Theta|\Theta_i^{k-1})$)
    \STATE $\bar w_i^{k} \leftarrow w_i^{k-1}\, p(z_k|p_k,\Theta_i^{k})$
  \ENDFOR
  \STATE Normalize $w^k$
  \STATE Compute ESS; if low then resample ($w^k\leftarrow 1/N$)
  \STATE (ATT) $w^k \leftarrow (1-\varepsilon)w^k+\varepsilon (A^{(s)})^\top w^k$; normalize
  \STATE (MH) if resampled then rejuvenate particles
  \STATE Compute $\mu_k,\Sigma_k$; $\sigma_k\leftarrow \sqrt{\mathrm{diag}(\Sigma_k)}$
  \STATE $s_k \leftarrow [p_k,z_k,\mu_k,\sigma_k]$, sample $a_k\sim\pi_\theta(\cdot|s_k)$, move
\ENDWHILE

\STATE Output \(\mu_k\) (and \(\Sigma_k\)) as the estimated source parameters.
\end{algorithmic}
\end{algorithm}

% =========================
% Section 5 Experiments (all-in-main-text version)
% Replace your current Section 5 with the following.
% =========================
\newpage

\section{Experiments}
\label{sec:experiments}

\begin{table}[htbp]
  \centering
  \resizebox{\linewidth}{!}{%
  \begin{tabular}{llll}
    \toprule
    \textbf{Source parameter} & \textbf{Distribution} 
    & \textbf{Source parameter} & \textbf{Distribution} \\
    \midrule
    $x_s,y_s$ (location) & $U(5,20)$ 
    & Release strength $q_s$ & $U(10,3000)$ \\
    Wind velocity $(u_x,u_y)$ & $U(0,6)$
    & Decay parameter $\lambda$ & $U(0,8)$ \\
    Diffusivity $\alpha$ & $U(1,5)$
    & Wind direction $\varphi_s$ & $U(0,2\pi)$ \\
    \bottomrule
  \end{tabular}}
  \caption{Parameter distributions for training scenarios.}
  \label{tab:training_parameters_}
\end{table}

In this paper, we utilize the ISLC environments (ISLCenv) to study active source localization under sparse feedback.
ISLCenv \cite{shi2024reinforcement} uses a Gaussian model to simulate field distributions and a sensor model to produce intensity observations.
The environment provides only positional and intensity observations (no shaped rewards), stressing (i) belief inference under particle degeneracy and (ii) generalizable information-seeking policies.

% -------------------------
% 5.1 Setup
% -------------------------
\subsection{Setup: environment, baselines, and metrics}
\label{sec:exp_setup}

\paragraph{Scenario generation (ID vs. OOD).}
Training scenarios are defined in a $25 \times 25$ area and generated by randomly sampling source/environment parameters
at the start of each episode from Table~\ref{tab:training_parameters_}.
Unless otherwise stated, we set the ESS resampling threshold ratio $\eta=0.6$ (Sec.~4.2) and the default cessation threshold $\zeta=0.5$ (Sec.~4.3).
The agent starts uniformly in $(0,5)\times(0,5)$ and moves with step size 1.
We evaluate each setting on 1,000 unseen test scenarios.

For out-of-distribution (OOD) experiments, the training region is confined to $(10,15)\times(10,15)$,
while testing regions are $(5,10)\times(15,20)$ and $(15,20)\times(15,20)$ (Fig.~\ref{fig_ood_Experiments_}).
This spatially disjoint split evaluates whether policies rely on belief-conditioned decision rules rather than memorizing region-specific trajectories. \textbf{Fields.} We test seven field modalities: Temperature (Temp.), Concentration (Conc.), Magnetic (Mag.), Electric (Elec.), Gas (Gas), Energy (En.), and Noise (Noise), which differ in signal characteristics and noise patterns. \textbf{Baselines.}
We compare against two groups:
(i) RL+Bayes methods: AGDC-KLD \cite{shi2025autonomous} and its variants AGDC-ENT and AGDC-EE;
(ii) planning+Bayes methods: Infotaxis \cite{vergassola2007infotaxis}, Entrotaxis \cite{hutchinson2018entrotaxis}, and DCEE \cite{chen2021dual}.
A Random policy is included as a lower bound. \textbf{Metrics.}
We report: \textbf{(1) OCE} (Operational Completion Efficacy), the stop rate under each method’s stopping rule.
For contraction-based methods, $k_{\text{stop}}$ is the first step satisfying $\|\mathrm{STD}_{k_{\text{stop}}}\|_\infty < \zeta$ (Sec.~4.3),
otherwise $k_{\text{stop}}=K_{\max}$, and $\mathrm{OCE}=\Pr[k_{\text{stop}}<K_{\max}]$.
\textbf{(2) ADE} (Average Deployment Efficiency), the traveled path length up to $k_{\text{stop}}$,
$\mathrm{ADE}=\mathbb{E}\big[\sum_{k=0}^{k_{\text{stop}}-1}\|p_{k+1}-p_k\|_2\big]$.
\textbf{(3) LPS} (Localization Precision Score), the terminal localization error of the posterior mean,
$\mathrm{LPS}=\mathbb{E}\big[\|(\mu^x_{k_{\text{stop}}},\mu^y_{k_{\text{stop}}})-(x^\ast,y^\ast)\|_2\big]$.
We additionally report \textbf{REV} (runtime proxy; lower is faster) for completeness, but emphasize OCE/ADE/LPS for task performance.

% -------------------------
% 5.2 Main results (ID)
% -------------------------
\subsection{Main results: in-distribution performance}
\label{sec:main_id_results}

\begin{table}[!h]
\centering
\resizebox{\linewidth}{!}{%
\begin{tabular}{lccccccc}
\toprule
\textbf{Method} & \textbf{Temp.} & \textbf{Conc.} & \textbf{Mag.} & \textbf{Elec.} & \textbf{Gas} & \textbf{En.} & \textbf{Noise} \\
\midrule

\multicolumn{8}{c}{\textbf{OCE}}\\
\midrule
ATT-PFRL     & 0.95$\pm$0.05 & 0.94$\pm$0.05 & 0.94$\pm$0.05 & 0.82$\pm$0.04 & 0.96$\pm$0.05 & 0.63$\pm$0.03 & 0.94$\pm$0.05 \\
AGDC-KLD       & 0.90$\pm$0.05 & 0.91$\pm$0.05 & 0.89$\pm$0.04 & 0.77$\pm$0.04 & 0.92$\pm$0.05 & 0.61$\pm$0.03 & 0.91$\pm$0.05 \\
AGDC-ENT        & 0.80$\pm$0.04 & 0.81$\pm$0.04 & 0.81$\pm$0.04 & 0.68$\pm$0.03 & 0.79$\pm$0.04 & 0.51$\pm$0.03 & 0.80$\pm$0.04 \\
AGDC-EE        & 0.87$\pm$0.04 & 0.88$\pm$0.04 & 0.86$\pm$0.04 & 0.74$\pm$0.04 & 0.86$\pm$0.04 & 0.57$\pm$0.03 & 0.86$\pm$0.04 \\
Infotaxis & 0.85$\pm$0.04 & 0.86$\pm$0.04 & 0.85$\pm$0.04 & 0.75$\pm$0.04 & 0.84$\pm$0.04 & 0.55$\pm$0.03 & 0.80$\pm$0.04 \\
Entrotaxis& 0.24$\pm$0.01 & 0.23$\pm$0.01 & 0.25$\pm$0.01 & 0.15$\pm$0.01 & 0.22$\pm$0.01 & 0.14$\pm$0.01 & 0.23$\pm$0.01 \\
DCEE     & 0.58$\pm$0.03 & 0.59$\pm$0.03 & 0.58$\pm$0.03 & 0.43$\pm$0.02 & 0.56$\pm$0.03 & 0.36$\pm$0.02 & 0.57$\pm$0.03 \\
Random              & $<$0.05       & $<$0.05       & $<$0.05       & $<$0.05       & $<$0.05       & $<$0.05       & $<$0.05       \\

\midrule
\multicolumn{8}{c}{\textbf{ADE}}\\
\midrule
ATT-PFRL     & 20$\pm$1.0 & 19$\pm$1.0 & 18$\pm$0.9 & 19$\pm$0.8 & 17$\pm$0.9 & 19$\pm$0.5 & 19$\pm$1.0 \\
AGDC-KLD      & 23$\pm$1.2 & 22$\pm$1.1 & 22$\pm$1.1 & 23$\pm$0.9 & 20$\pm$1.0 & 22$\pm$0.6 & 21$\pm$1.1 \\
AGDC-ENT       & 25$\pm$1.3 & 24$\pm$1.2 & 25$\pm$1.2 & 20$\pm$1.0 & 22$\pm$1.1 & 24$\pm$0.7 & 23$\pm$1.2 \\
AGDC-EE       & 25$\pm$1.3 & 24$\pm$1.2 & 24$\pm$1.2 & 19$\pm$1.0 & 21$\pm$1.1 & 23$\pm$0.7 & 22$\pm$1.1 \\
Infotaxis & 50$\pm$2.5 & 48$\pm$2.4 & 51$\pm$2.4 & 58$\pm$1.9 & 43$\pm$2.2 & 47$\pm$1.4 & 45$\pm$2.3 \\
Entrotaxis& 62$\pm$3.1 & 60$\pm$3.0 & 59$\pm$3.0 & 61$\pm$2.5 & 56$\pm$2.8 & 55$\pm$1.8 & 58$\pm$2.9 \\
DCEE      & 57$\pm$2.9 & 55$\pm$2.8 & 54$\pm$2.7 & 55$\pm$2.3 & 51$\pm$2.6 & 57$\pm$1.6 & 53$\pm$2.7 \\
Random              & $>$150     & $>$150     & $>$150     & $>$150     & $>$150     & $>$150     & $>$150     \\

\midrule
\multicolumn{8}{c}{\textbf{REV}}\\
\midrule
ATT-PFRL     & 0.15$\pm$0.08 & 0.14$\pm$0.07 & 0.14$\pm$0.07 & 0.12$\pm$0.06 & 0.14$\pm$0.07 & 0.09$\pm$0.05 & 0.13$\pm$0.07 \\
AGDC-KLD       & 0.10$\pm$0.05 & 0.10$\pm$0.05 & 0.10$\pm$0.05 & 0.09$\pm$0.05 & 0.10$\pm$0.05 & 0.07$\pm$0.04 & 0.10$\pm$0.05 \\
AGDC-ENT      & 0.10$\pm$0.05 & 0.10$\pm$0.05 & 0.10$\pm$0.05 & 0.09$\pm$0.05 & 0.10$\pm$0.05 & 0.07$\pm$0.04 & 0.10$\pm$0.05 \\
AGDC-EE        & 0.10$\pm$0.05 & 0.10$\pm$0.05 & 0.10$\pm$0.05 & 0.09$\pm$0.05 & 0.10$\pm$0.05 & 0.07$\pm$0.04 & 0.10$\pm$0.05 \\
Infotaxis & 1.50$\pm$0.08 & 1.40$\pm$0.07 & 1.40$\pm$0.07 & 1.20$\pm$0.06 & 1.30$\pm$0.07 & 0.80$\pm$0.04 & 1.30$\pm$0.07 \\
Entrotaxis& 1.40$\pm$0.07 & 1.30$\pm$0.07 & 1.30$\pm$0.07 & 1.10$\pm$0.06 & 1.20$\pm$0.06 & 0.70$\pm$0.04 & 1.20$\pm$0.06 \\
DCEE      & 1.40$\pm$0.07 & 1.30$\pm$0.07 & 1.30$\pm$0.07 & 1.10$\pm$0.06 & 1.20$\pm$0.06 & 0.70$\pm$0.04 & 1.20$\pm$0.06 \\
Random              & $<$0.01       & $<$0.01       & $<$0.01       & $<$0.01       & $<$0.01       & $<$0.01       & $<$0.01       \\

\midrule
\multicolumn{8}{c}{\textbf{LPS}}\\
\midrule
ATT-PFRL      & 0.05$\pm$0.01 & 0.05$\pm$0.01 & 0.05$\pm$0.01 & 0.08$\pm$0.01 & 0.05$\pm$0.01 & 0.06$\pm$0.01 & 0.05$\pm$0.01 \\
AGDC-KLD       & 0.20$\pm$0.01 & 0.20$\pm$0.01 & 0.20$\pm$0.01 & 0.17$\pm$0.01 & 0.20$\pm$0.01 & 0.13$\pm$0.01 & 0.20$\pm$0.01 \\
AGDC-ENT       & 0.25$\pm$0.01 & 0.24$\pm$0.01 & 0.24$\pm$0.01 & 0.20$\pm$0.01 & 0.22$\pm$0.01 & 0.14$\pm$0.01 & 0.23$\pm$0.01 \\
AGDC-EE        & 0.23$\pm$0.01 & 0.22$\pm$0.01 & 0.22$\pm$0.01 & 0.18$\pm$0.01 & 0.20$\pm$0.01 & 0.12$\pm$0.01 & 0.21$\pm$0.01 \\
Infotaxis & 0.60$\pm$0.03 & 0.60$\pm$0.03 & 0.60$\pm$0.03 & 0.51$\pm$0.02 & 0.60$\pm$0.03 & 0.39$\pm$0.02 & 0.60$\pm$0.03 \\
Entrotaxis& 0.70$\pm$0.04 & 0.70$\pm$0.04 & 0.70$\pm$0.04 & 0.60$\pm$0.03 & 0.70$\pm$0.04 & 0.46$\pm$0.02 & 0.70$\pm$0.04 \\
DCEE     & 0.60$\pm$0.03 & 0.60$\pm$0.03 & 0.60$\pm$0.03 & 0.51$\pm$0.02 & 0.60$\pm$0.03 & 0.39$\pm$0.02 & 0.60$\pm$0.03 \\
Random              & 5.0$\pm$0.25  & 5.0$\pm$0.25  & 5.0$\pm$0.25  & 5.0$\pm$0.25  & 5.0$\pm$0.25  & 5.0$\pm$0.25  & 5.0$\pm$0.25  \\
\bottomrule
\end{tabular}}
\caption{Comparison under in-distribution (ID) scenarios.}
\label{tab:fundamental_experiments_}
\end{table}

Table~\ref{tab:fundamental_experiments_} provides end-to-end evidence that ATT-PFRL improves both task completion and localization accuracy across all field modalities.
High OCE indicates that the learned policy can reliably trigger termination, while low LPS indicates that termination typically occurs near the true source rather than due to accidental contraction.The large ADE gap vs. planning baselines suggests that hand-crafted heuristics can be myopic under noise/model mismatch, while belief-conditioned RL learns more efficient information gathering.

% -------------------------
% 5.3 OOD generalization
% -------------------------
\subsection{Generalization: out-of-distribution}
\label{sec:ood_results}

\begin{figure}[htbp]
  \centering
  \begin{subfigure}[t]{0.47\linewidth}
    \includegraphics[width=\linewidth]{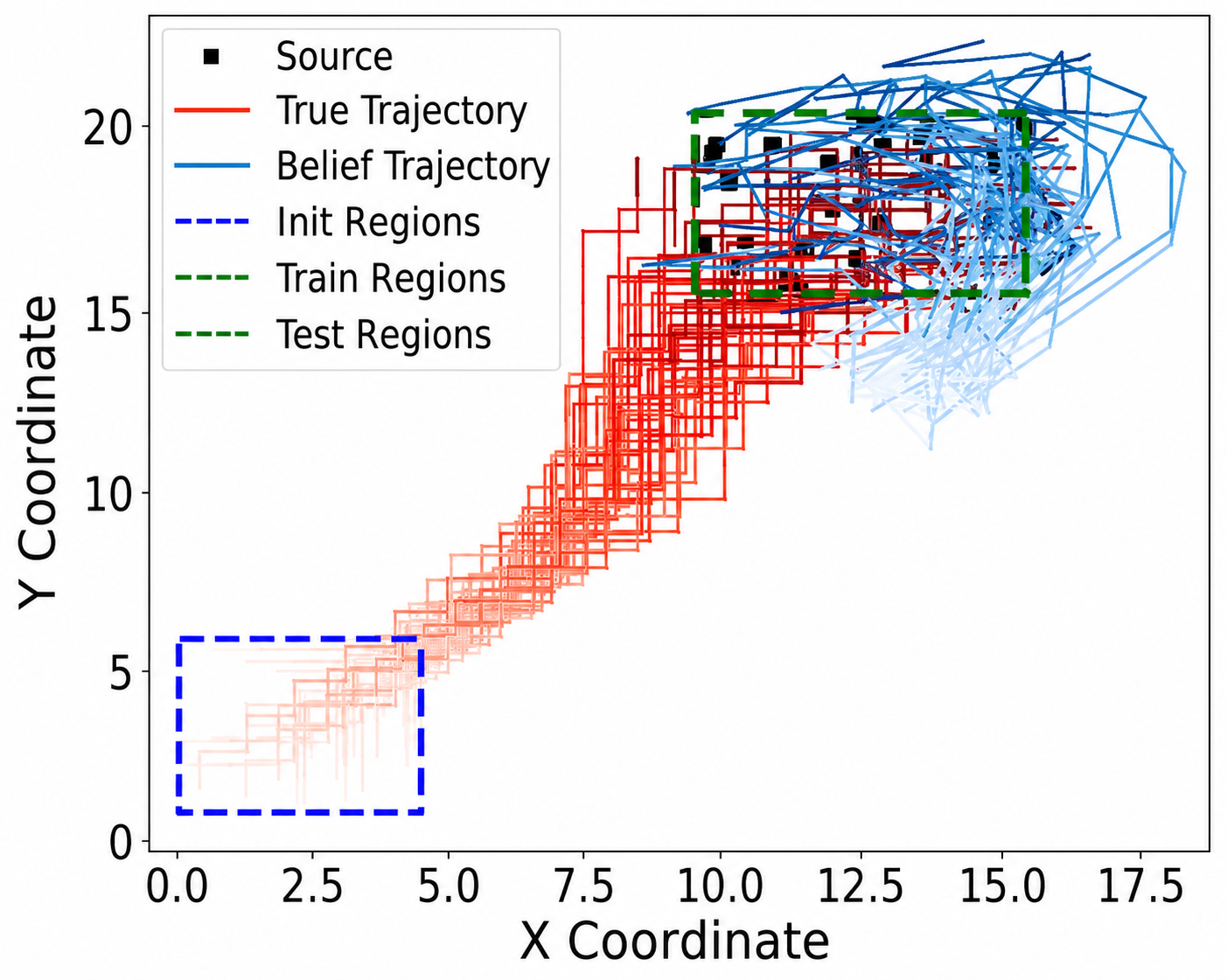}
    \caption{In-distribution (ID)}
    \label{fig_Fundamental_Experiments}
  \end{subfigure}
  \begin{subfigure}[t]{0.47\linewidth}
    \includegraphics[width=\linewidth]{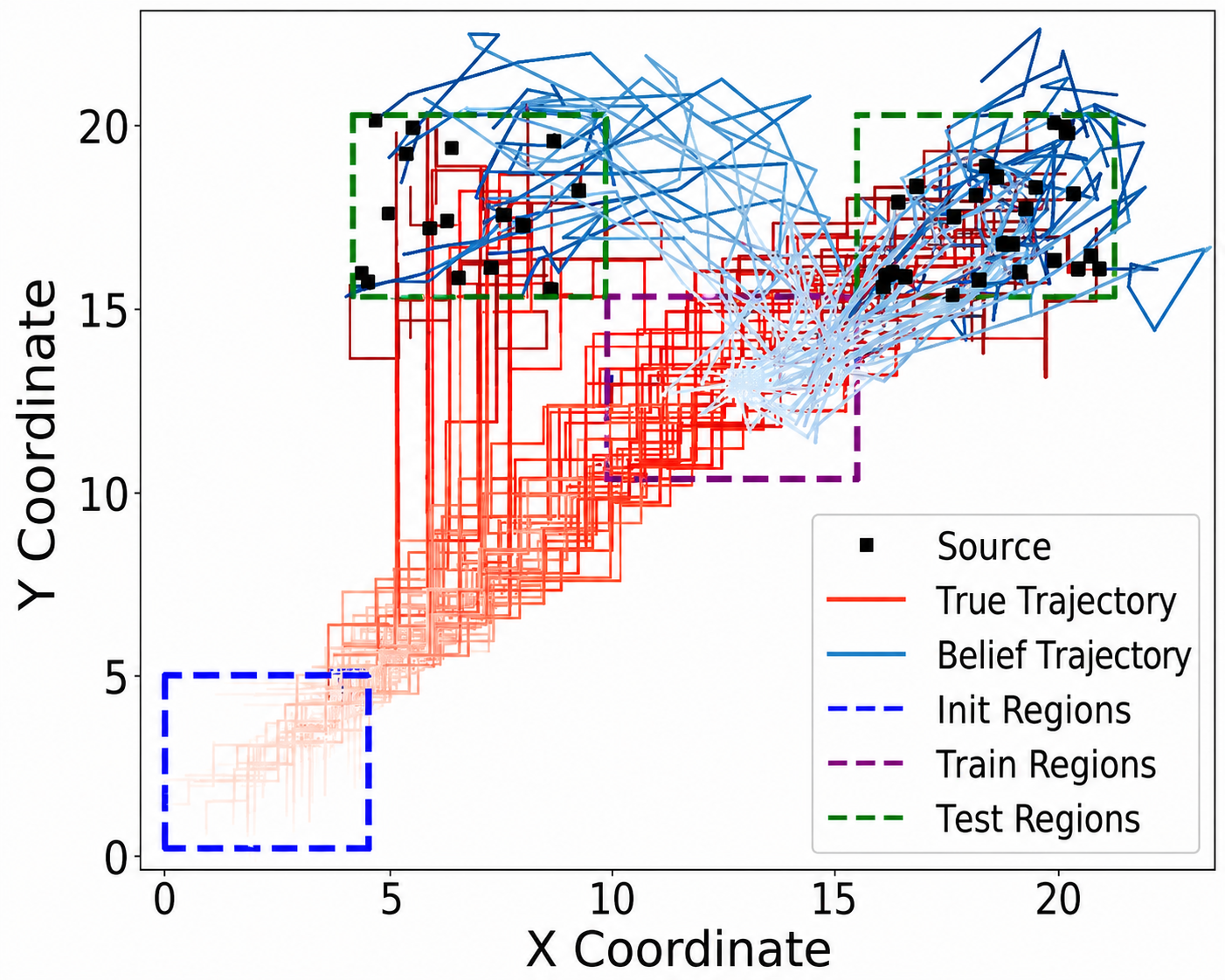}
    \caption{Out-of-distribution (OOD)}
    \label{fig_ood_Experiments_}
  \end{subfigure}
  \caption{Visualization of test trajectories (100 episodes).}
  \label{fig_id_ood_traj}
\end{figure}

\begin{table}[htbp]
\centering
\setlength{\tabcolsep}{4pt}
\renewcommand{\arraystretch}{0.95}
\resizebox{\linewidth}{!}{
\begin{tabular}{lccccccc}
\toprule
\textbf{Method} & \textbf{Temp.} & \textbf{Conc.} & \textbf{Mag.} & \textbf{Elec.} & \textbf{Gas} & \textbf{En.} & \textbf{Noise} \\
\midrule
ATT-PFRL (ours) & 0.95 & 0.94 & 0.94 & 0.82 & 0.96 & 0.63 & 0.94 \\
AGDC-KLD        & 0.448 & 0.453 & 0.446 & 0.382 & 0.458 & 0.307 & 0.452 \\
AGDC-ENT        & 0.271 & 0.268 & 0.273 & 0.225 & 0.265 & 0.168 & 0.269 \\
AGDC-EE         & 0.288 & 0.291 & 0.285 & 0.249 & 0.289 & 0.192 & 0.285 \\
\bottomrule
\end{tabular}}
\caption{OOD generalization across fields (spatially shifted regions)}
\label{tab:ood_compact}
\end{table}

\noindent OOD performance drops substantially for RL+Bayes baselines, while ATT-PFRL maintains high completion rates across fields (Table~\ref{tab:ood_compact}).
Fig.~\ref{fig_ood_Experiments_} further shows that belief trajectories continue to converge near the source under spatial shifts, providing a process-level explanation for the aggregated OCE results.
Since OOD amplifies inference errors, we next test inference robustness via fixed-trajectory evaluation.

% -------------------------
% 5.3b Attention ablation + diagnostics
% -------------------------
\subsection{Ablation on feature-aware attention}
\label{sec:att_ablation}

\begin{figure}[!h]
\centering
\includegraphics[width=\linewidth,height=4cm]{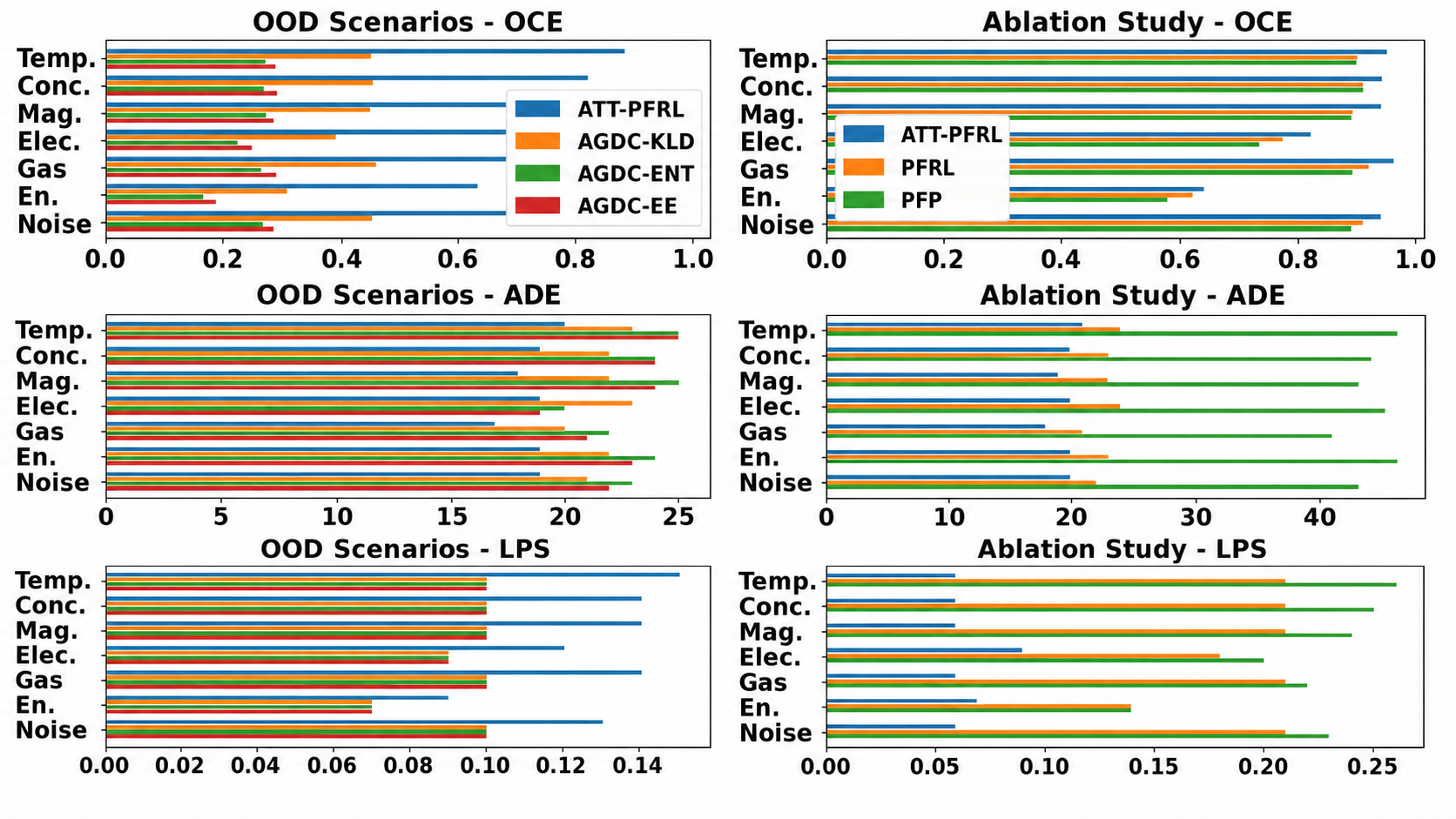}
\caption{OOD robustness (left) and attention ablation (right).}
\label{fig_ood_ablation}
\end{figure}

\begin{figure}[htbp]
  \centering
  \begin{subfigure}[t]{0.45\linewidth}
    \includegraphics[height=3cm]{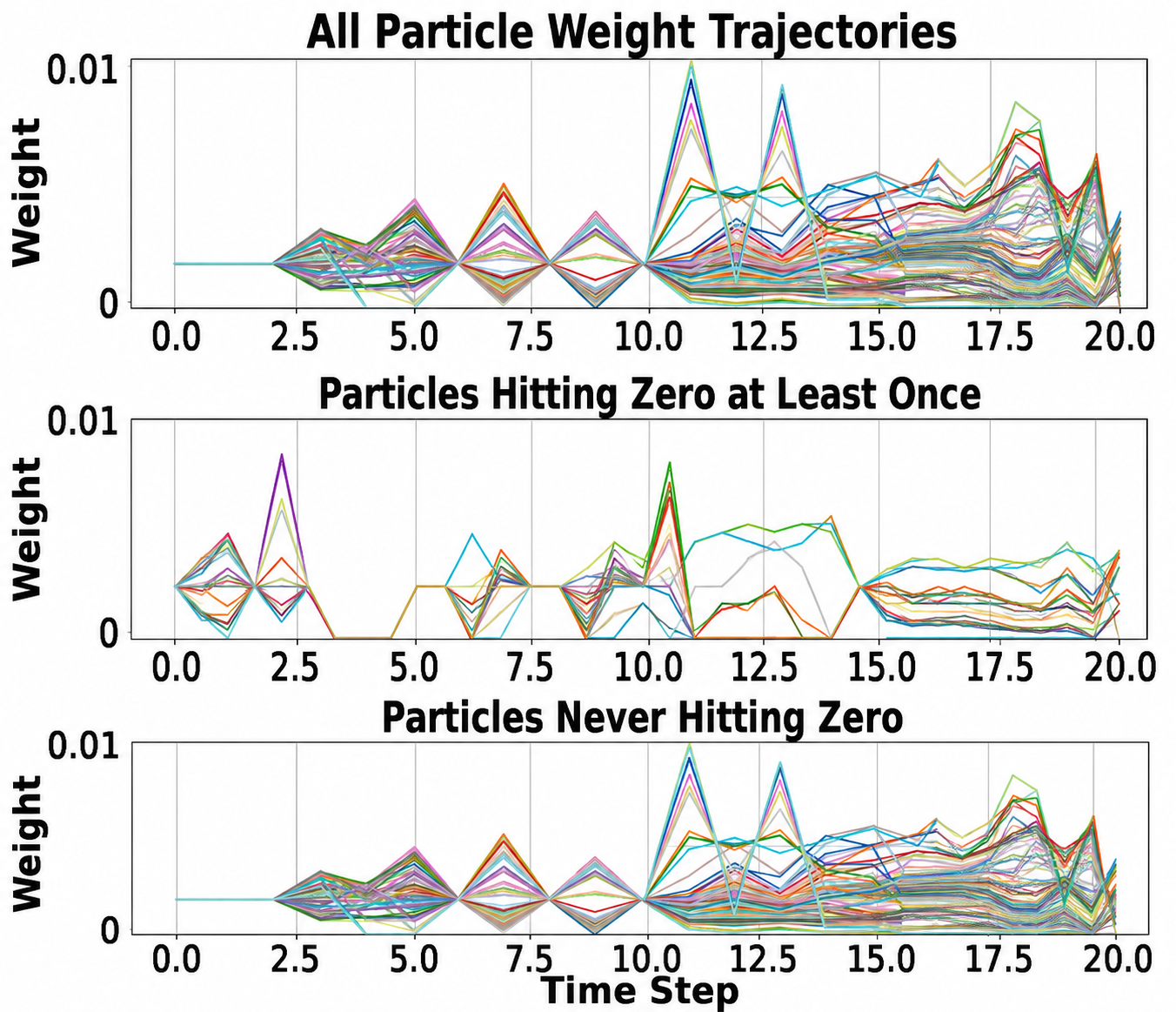}
    \caption{$w$ with attention}
    \label{fig_plot_with_att}
  \end{subfigure}
  \begin{subfigure}[t]{0.45\linewidth}
    \includegraphics[height=3cm]{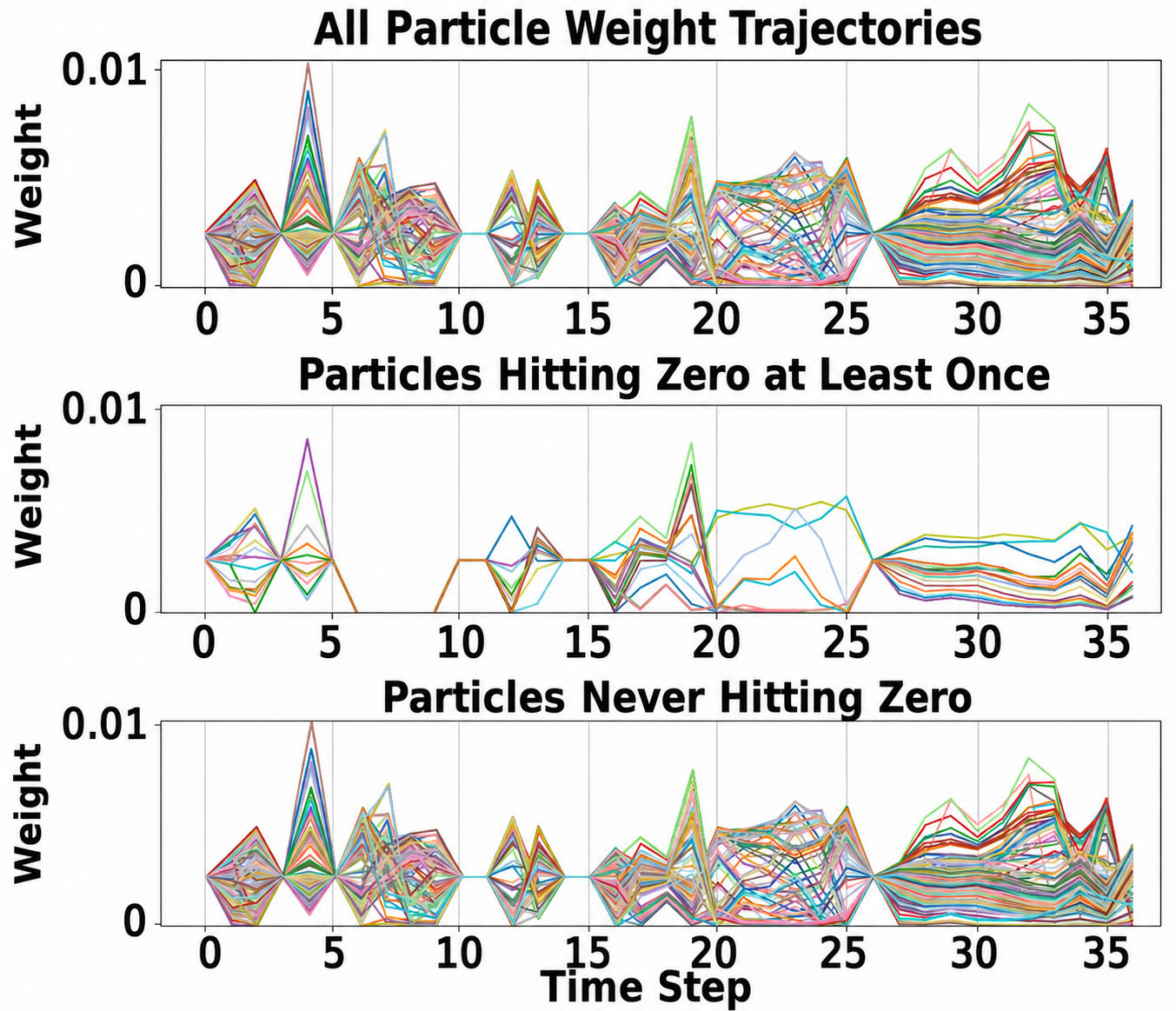}
    \caption{$w$ without attention}
    \label{fig_plot_without_att}
  \end{subfigure}
  \caption{Particle-weight evolution under identical conditions.}
  \label{fig_weight_evolution}
\end{figure}

Fig.~\ref{fig_ood_ablation} (right) shows that removing attention (PFRL) consistently degrades both completion and localization quality across fields,
demonstrating that attention is not a cosmetic modification but a core component for robust belief updates.
The weight-evolution diagnostics (Fig.~\ref{fig_weight_evolution}) explain this gap: without attention, particle weights more frequently collapse to zero,
reducing effective sample size and making posterior estimates brittle.
This qualitative mechanism-level evidence complements the quantitative inference decoupling results in Table~\ref{tab:fixed_trajectory_inference_},
where attention increases ESS and reduces RMSE under controlled observation sequences.

% -------------------------
% 5.3c Uniform mixing control + prior robustness
% -------------------------
\subsection{Uniform mixing lacks attention}
\label{sec:uniform_mixing}

Uniform mixing is a special case of convex smoothing on the weight simplex: $\tilde{w}_i=(1-\rho)w_i+\rho\frac{1}{N}, \rho\in[0,1]$, which corresponds to replacing feature-aware attention with a uniform row-stochastic matrix.
We test whether the gain of attention is merely due to generic smoothing by sweeping $\rho$.

\begin{table}[htbp]
\centering
\resizebox{\linewidth}{!}{
\begin{tabular}{ccccc}
\hline
$\rho$ & ADE (Basic) & OCE (Basic) & ADE (OOD) & OCE (OOD) \\
\hline
0.0 & 22$\pm$1.1 & 0.93$\pm$0.05 & 28$\pm$1.4 & 0.88$\pm$0.04 \\
0.1 & 19$\pm$1.0 & 0.95$\pm$0.05 & 24$\pm$1.2 & 0.91$\pm$0.05 \\
0.2 & 18$\pm$0.9 & 0.96$\pm$0.05 & 23$\pm$1.2 & 0.92$\pm$0.05 \\
0.3 & 20$\pm$1.0 & 0.94$\pm$0.05 & 26$\pm$1.3 & 0.89$\pm$0.04 \\
0.4 & 25$\pm$1.3 & 0.90$\pm$0.05 & 32$\pm$1.6 & 0.84$\pm$0.04 \\
0.5 & 32$\pm$1.6 & 0.85$\pm$0.04 & 40$\pm$2.0 & 0.78$\pm$0.04 \\
\hline
\end{tabular}}
\caption{Sensitivity to $\rho$ in uniform mixing (convex smoothing).}
\label{tab:uniform_mixing_comparison_}
\end{table}

\begin{table}[htbp]
\centering
\setlength{\tabcolsep}{5pt}
\renewcommand{\arraystretch}{0.95}
\resizebox{\linewidth}{!}{
\begin{tabular}{lccc lccc}
\toprule
\multicolumn{4}{c}{\textbf{Convex}} & \multicolumn{4}{c}{\textbf{Non-convex}} \\
\cmidrule(lr){1-4}\cmidrule(lr){5-8}
Dist. & OCE & ADE & LPS & Dist. & OCE & ADE & LPS \\
\midrule
Beta      & 0.95$\pm$0.05 & 19$\pm$1.0 & 0.05$\pm$0.01 & Star Shape   & 0.92$\pm$0.05 & 22$\pm$1.1 & 0.07$\pm$0.01 \\
Gaussian  & 0.96$\pm$0.05 & 18$\pm$0.9 & 0.04$\pm$0.01 & Quarter Ring & 0.90$\pm$0.05 & 24$\pm$1.2 & 0.08$\pm$0.01 \\
Dirichlet & 0.94$\pm$0.05 & 20$\pm$1.0 & 0.06$\pm$0.01 & Half Ring    & 0.88$\pm$0.04 & 26$\pm$1.3 & 0.09$\pm$0.01 \\
Uniform   & 0.95$\pm$0.05 & 19$\pm$1.0 & 0.05$\pm$0.01 & Full Ring    & 0.86$\pm$0.04 & 28$\pm$1.4 & 0.10$\pm$0.01 \\
\bottomrule
\end{tabular}}
\caption{Robustness to prior mismatch in the initial belief.}
\label{tab:complex_distribution_}
\end{table}

Table~\ref{tab:uniform_mixing_comparison_} shows a sweet-spot behavior: mild uniform mixing ($\rho\approx 0.1$--$0.2$) slightly improves OCE/ADE by mitigating extreme collapse,
but larger $\rho$ biases the posterior toward uniformity and degrades both ID and OOD performance.
This control rules out the alternative explanation that ``any smoothing'' would work; the benefit comes from feature-aware mass transfer that preserves informative modes.
Table~\ref{tab:complex_distribution_} further demonstrates robustness to prior-shape mismatch (including non-convex priors),
supporting that attention regularization does not rely on a fortunate prior geometry and remains stable under diverse belief initializations.

% -------------------------
% 5.3d Nonstationary shift robustness
% -------------------------
\subsection{Adaptation under nonstationary source shifts}
\label{sec:nonstationary}

We evaluate adaptability in a nonstationary setting where the source location undergoes an abrupt change during inference:
after 15 steps, the source position shifts to a new location while the agent continues operating without reset.
This tests whether the inference layer can rapidly reallocate posterior mass after a regime change.

\begin{table}[htbp]
\centering
\resizebox{\linewidth}{!}{
\begin{tabular}{lcccccc}
\hline
Field & \multicolumn{2}{c}{OCE} & \multicolumn{2}{c}{ADE} & \multicolumn{2}{c}{LPS} \\
 & Ours & AGDC-KLD & Ours & AGDC-KLD & Ours & AGDC-KLD \\
\hline
Temperature & \textbf{0.95$\pm$0.05} & 0.90$\pm$0.05 & \textbf{20$\pm$1.0} & 23$\pm$1.2 & \textbf{0.05$\pm$0.01} & 0.20$\pm$0.01 \\
Concentration & \textbf{0.94$\pm$0.05} & 0.91$\pm$0.05 & \textbf{19$\pm$1.0} & 22$\pm$1.1 & \textbf{0.05$\pm$0.01} & 0.20$\pm$0.01 \\
Magnetic & \textbf{0.94$\pm$0.05} & 0.89$\pm$0.04 & \textbf{18$\pm$0.9} & 22$\pm$1.1 & \textbf{0.05$\pm$0.01} & 0.20$\pm$0.01 \\
Electric & \textbf{0.82$\pm$0.04} & 0.77$\pm$0.04 & \textbf{19$\pm$0.8} & 23$\pm$0.9 & \textbf{0.08$\pm$0.01} & 0.17$\pm$0.01 \\
Gas & \textbf{0.96$\pm$0.05} & 0.92$\pm$0.05 & \textbf{17$\pm$0.9} & 20$\pm$1.0 & \textbf{0.05$\pm$0.01} & 0.20$\pm$0.01 \\
Energy & \textbf{0.63$\pm$0.03} & 0.61$\pm$0.03 & \textbf{19$\pm$0.5} & 22$\pm$0.6 & \textbf{0.06$\pm$0.01} & 0.13$\pm$0.01 \\
Noise & \textbf{0.94$\pm$0.05} & 0.91$\pm$0.05 & \textbf{19$\pm$1.0} & 21$\pm$1.1 & \textbf{0.05$\pm$0.01} & 0.20$\pm$0.01 \\
\hline
\end{tabular}}
\caption{Adaptability to nonstationary conditions.}
\label{tab:nonstationary_conditions_}
\end{table}

Nonstationarity is a stricter stress test than spatial OOD: the agent must abandon an already contracting belief and re-localize without resetting exploration.
ATT-PFRL maintains higher completion and lower localization error than AGDC-KLD, suggesting that degeneracy control and rejuvenation help preserve particle diversity,
which in turn enables faster posterior reallocation after regime changes.
This observation is consistent with the fixed-trajectory results in Sec.~\ref{sec:inference_layer_decoupling}: higher ESS correlates with stronger recovery from weak or shifting observations in Table \ref{tab:nonstationary_conditions_}.

% -------------------------
% 5.4 Inference ablation
% -------------------------
\subsection{Inference ablation: fixed-trajectory evaluation}
\label{sec:inference_layer_decoupling}

\begin{table}[htbp]
\centering
\setlength{\tabcolsep}{6pt}
\renewcommand{\arraystretch}{1.0}
\resizebox{\linewidth}{!}{
\begin{tabular}{lcccc}
\toprule
Method & RMSE@20 (m)$\downarrow$ & ESS@20$\uparrow$ & Likelihood Evals$\downarrow$ & Wall-time (ms)$\downarrow$ \\
\midrule
PF-baseline & 5.2$\pm$0.8 & 120$\pm$35 & 10000 & 45$\pm$8 \\
PF+Attention ($R{=}0$) & 3.6$\pm$0.6 & 240$\pm$50 & 10500 & 52$\pm$9 \\
PF+MH ($\varepsilon{=}0$) & 4.1$\pm$0.7 & 180$\pm$42 & 15000 & 68$\pm$12 \\
\textbf{PF+Att+MH (Ours)} & \textbf{2.8$\pm$0.4} & \textbf{285$\pm$48} & 15800 & 72$\pm$11 \\
\bottomrule
\end{tabular}}
\caption{Inference quality on fixed random-walk trajectories.}
\label{tab:fixed_trajectory_inference_}
\end{table}

To isolate the contribution of the \emph{inference layer} (ESS-triggered resampling, feature-aware attention smoothing, and MH rejuvenation)
from policy-induced data collection, we compare inference variants under identical observation sequences $o_{1:K}$.
We evaluate on the most challenging setting: fixed random-walk trajectories where observations are weakly informative. Table~\ref{tab:fixed_trajectory_inference_} shows that attention smoothing substantially increases ESS and reduces RMSE, indicating that it directly mitigates weight collapse. MH rejuvenation alone yields a smaller gain, while combining attention and MH achieves the best RMSE/ESS,
supporting the complementarity claimed in Sec.~4.2: attention prevents extreme collapse, and MH restores diversity after resampling.
This provides a causal explanation for the end-to-end improvements in Table~\ref{tab:fundamental_experiments_} and the robustness observed under OOD tests.

% -------------------------
% 5.5 Cessation & reward validation
% -------------------------
\subsection{Cessation \& reward validation}
\label{sec:termination_calibration_reward}

Our framework uses belief contraction both as a termination criterion and as an intrinsic learning signal:
termination occurs when $\|\mathrm{STD}_k\|_\infty < \zeta$, and
$r_{k+1}=\mathbb{I}[\|\mathrm{STD}_{k+1}\|_\infty < \zeta]$.
This subsection validates (i) threshold sensitivity, (ii) calibration against ground-truth accuracy, and (iii) reward design.

\textbf{Threshold sensitivity with GT-success.}
To avoid circular evaluation (defining success using the same threshold being swept), we evaluate termination using an independent ground-truth criterion:
let $e_{\mathrm{stop}}=\|(\mu_x,\mu_y)-(x^\star,y^\star)\|_2$ at termination and define GT-success as $e_{\mathrm{stop}}<\epsilon_{\mathrm{loc}}$ with $\epsilon_{\mathrm{loc}}=1.0$m.
We sweep $\zeta$ only at test time, keeping policy and inference fixed.

\begin{table}[htbp]
\centering
\setlength{\tabcolsep}{5pt}
\renewcommand{\arraystretch}{0.95}
\resizebox{\linewidth}{!}{
\begin{tabular}{cccccc}
\toprule
$\zeta$ & Stop Rate $\uparrow$ & GT-Success $\uparrow$ & Avg.~Steps $\downarrow$ & ADE $\downarrow$ & False Stop $\downarrow$ \\
\midrule
0.2 & 0.88 & \textbf{0.96} & 31$\pm$6 & 22$\pm$2 & 0.02 \\
0.3 & 0.92 & \textbf{0.95} & 28$\pm$5 & 21$\pm$2 & 0.04 \\
0.4 & 0.94 & 0.94 & 24$\pm$4 & 20$\pm$1 & 0.06 \\
\textbf{0.5} & \textbf{0.95} & 0.92 & \textbf{22$\pm$4} & \textbf{20$\pm$1} & \textbf{0.08} \\
0.6 & 0.96 & 0.89 & 20$\pm$3 & 19$\pm$1 & 0.12 \\
0.7 & 0.97 & 0.86 & 18$\pm$3 & 19$\pm$2 & 0.15 \\
0.8 & 0.98 & 0.81 & 16$\pm$3 & 18$\pm$2 & 0.21 \\
1.0 & 0.99 & 0.75 & 14$\pm$2 & 18$\pm$2 & 0.28 \\
\bottomrule
\end{tabular}}
\caption{Cessation threshold sweep on Temperature (ID).}
\label{tab:threshold_sensitivity_basic_}
\end{table}

\begin{table}[htbp]
\centering
\setlength{\tabcolsep}{6pt}
\renewcommand{\arraystretch}{0.95}
\resizebox{0.9\linewidth}{!}{
\begin{tabular}{lccccc}
\toprule
Setting & Stop Rate $\uparrow$ & GT-Success $\uparrow$ & Avg.~Steps $\downarrow$ & ADE $\downarrow$ & False Stop $\downarrow$ \\
\midrule
Basic (ID) & 0.95 & 0.92 & 22$\pm$4 & 20$\pm$1 & 0.08 \\
OOD & 0.91 & 0.85 & 26$\pm$5 & 23$\pm$2 & 0.11 \\
Nonstationary & 0.93 & 0.82 & 24$\pm$5 & 21$\pm$2 & 0.14 \\
\bottomrule
\end{tabular}}
\caption{Robustness of default cessation threshold $\zeta=0.5$.}
\label{tab:threshold_robustness_}
\end{table}

As $\zeta$ increases, Stop Rate rises monotonically, but GT-Success decreases and False Stop increases, revealing a clear efficiency--accuracy trade-off.
The default $\zeta=0.5$ is a balanced operating point: it achieves strong GT-Success while avoiding excessive exploration (Table~\ref{tab:threshold_sensitivity_basic_}),
and remains stable under distribution shifts and nonstationarity (Table~\ref{tab:threshold_robustness_}).
This directly supports the claim that our stopping signal is not only effective (high OCE) but also controllable and transferable.
\begin{table}[htbp]
\centering
\setlength{\tabcolsep}{6pt}
\renewcommand{\arraystretch}{0.95}
\resizebox{0.95\linewidth}{!}{
\begin{tabular}{lcccc}
\toprule
Field & Spearman $\rho_{\mathrm{S}}$ & False Term. $\downarrow$ & Precision@$\zeta{=}0.5$ $\uparrow$ & Recall@$\zeta{=}0.5$ $\uparrow$ \\
\midrule
Temperature    & $-0.78$ & 0.08 & 0.92 & 0.89 \\
Concentration  & $-0.76$ & 0.09 & 0.91 & 0.87 \\
Magnetic       & $-0.80$ & 0.07 & 0.93 & 0.90 \\
Electric       & $-0.72$ & 0.12 & 0.88 & 0.84 \\
Gas            & $-0.81$ & 0.06 & 0.94 & 0.91 \\
Energy         & $-0.69$ & 0.15 & 0.85 & 0.80 \\
Noise          & $-0.77$ & 0.09 & 0.91 & 0.88 \\
\bottomrule
\end{tabular}}
\caption{Calibration of contraction $c_k$ vs. ground-truth error.}
\label{tab:calibration_metrics_}
\end{table}

\textbf{Calibration against ground-truth error.}
On 1,000 test episodes per field, we record contraction $c_k=\|\mathrm{STD}_k\|_\infty$ and ground-truth error $e_k$ over time.
We report Spearman rank correlation, false termination rate at $\zeta=0.5$, and precision/recall of $\mathbb{I}[c_k<\zeta]$.
The contraction signal is consistently and strongly negatively correlated with true localization error across all fields,
indicating that posterior contraction is a reliable proxy for accuracy rather than an arbitrary confidence measure.
Importantly, this calibration becomes meaningful in light of Sec.~\ref{sec:inference_layer_decoupling}:
since the inference layer demonstrably improves RMSE/ESS under controlled observations, the contraction signal is grounded in improved posterior quality, mitigating ``confident but wrong'' failure modes.

\begin{table}[htbp]
\centering
\setlength{\tabcolsep}{6pt}
\renewcommand{\arraystretch}{0.95}
\resizebox{0.9\linewidth}{!}{
\begin{tabular}{lcccc}
\toprule
Criterion & GT-Success $\uparrow$ & Avg.~Steps $\downarrow$ & False Term. $\downarrow$ & $\rho_{\mathrm{S}}$ $\downarrow$ \\
\midrule
$\|\mathrm{STD}\|_\infty < 0.5$ (ours) & \textbf{0.92} & 22$\pm$4 & \textbf{0.08} & \textbf{-0.78} \\
Entropy $<\tau_H$ & 0.87 & 24$\pm$5 & 0.14 & -0.68 \\
$\max_i w_i > \tau_w$ & 0.84 & 20$\pm$4 & 0.18 & -0.61 \\
Fixed steps ($k{=}25$) & 0.79 & 25 & 0.23 & N/A \\
\bottomrule
\end{tabular}}
\caption{Comparison of termination criteria (Temperature, ID).}
\label{tab:termination_criteria_comparison_}
\end{table}

We compare contraction-based cessation with entropy thresholding, max-weight thresholding, and fixed-step stopping on Temperature (ID) in Table \ref{tab:termination_criteria_comparison_}. This comparison shows that contraction-based cessation achieves the best overall balance: it yields higher GT-Success and lower premature termination than entropy/max-weight criteria.

\begin{table}[htbp]
\centering
\setlength{\tabcolsep}{5pt}
\renewcommand{\arraystretch}{0.95}
\resizebox{\linewidth}{!}{
\begin{tabular}{lccccc}
\toprule
Reward Type & OCE (ID) $\uparrow$ & GT-Success (ID) $\uparrow$ & ADE $\downarrow$ & OCE (OOD) $\uparrow$ & Sample Eff. $\downarrow$ \\
\midrule
R1: Indicator (ours) & \textbf{0.95} & \textbf{0.92} & \textbf{20$\pm$1} & \textbf{0.88} & 1200 eps to 0.90 \\
R2: Dense contraction & 0.91 & 0.87 & 21$\pm$1 & 0.84 & \textbf{980} eps to 0.90 \\
R3: Entropy reduction & 0.88 & 0.83 & 23$\pm$1 & 0.80 & 1450 eps to 0.90 \\
\bottomrule
\end{tabular}}
\caption{Reward design ablation for ATT-PFRL.}
\label{tab:reward_design_ablation_}
\end{table}

\noindent\textbf{Reward formulation ablation.} We train ATT-PFRL under three intrinsic reward variants with identical budgets: (R1) indicator (ours), (R2) dense contraction shaping, and (R3) entropy reduction in Table \ref{tab:reward_design_ablation_}. Dense shaping (R2/R3) can speed up early learning but degrades asymptotic performance and OOD generalization,
suggesting that optimizing a dense proxy (contraction reduction per step) can misalign the policy with the actual stopping objective.
In contrast, the indicator reward (R1) directly matches the cessation goal and therefore preserves goal alignment under shifts,
closing the loop between inference quality (Sec.~\ref{sec:inference_layer_decoupling}), calibrated stopping (Table~\ref{tab:calibration_metrics_}),
and robust end-to-end behavior (Secs.~\ref{sec:main_id_results}--\ref{sec:ood_results}).

\section{Conclusion}
We present ATT-PFRL for active source localization under sparse feedback by coupling belief inference with RL. 
The attention-augmented particle filter mitigates degeneracy and improves posterior quality. 
Experiments across diverse fields show higher completion and better localization than RL+Bayes and planning baselines, including OOD. 
Belief contraction is calibrated as a stopping signal and provides an intrinsic reward without hand-crafted distance shaping.

\clearpage
\section*{Acknowledgments}
The authors gratefully acknowledge financial support from a University of Bristol scholarship. This work was also supported in part by the Royal Society.

\bibliographystyle{named}
\bibliography{ijcai26}

\end{document}